\documentclass[10pt,twocolumn,letterpaper]{article}

\usepackage{cvpr}              

\usepackage{graphicx}
\usepackage{amsmath}
\usepackage{amssymb}
\usepackage{booktabs}
\usepackage{algorithm}  
\usepackage{algorithmicx}  
\usepackage{algpseudocode}
\usepackage{booktabs,tabularx}
\usepackage{multirow}

\usepackage[pagebackref,breaklinks,colorlinks]{hyperref}
\usepackage[accsupp]{axessibility}

\usepackage[capitalize]{cleveref}
\crefname{section}{Sec.}{Secs.}
\Crefname{section}{Section}{Sections}
\Crefname{table}{Table}{Tables}
\crefname{table}{Tab.}{Tabs.}

\def\cvprPaperID{4743} 
\def\confName{CVPR}
\def\confYear{2022}

\begin{document}

\title{Few Could Be Better Than All:

Feature Sampling and Grouping for Scene Text Detection}

\author{
    Jingqun Tang\textsuperscript{\rm 1},
    Wenqing Zhang\textsuperscript{\rm 2},
    Hongye Liu\textsuperscript{\rm 1},
    MingKun Yang\textsuperscript{\rm 2},\\
    Bo Jiang\textsuperscript{\rm 1},
    Guanglong Hu\textsuperscript{\rm 1},
    Xiang Bai\textsuperscript{\rm 2}\thanks{Corresponding Author}\\
    \textsuperscript{\rm 1}NetEase,
    \textsuperscript{\rm 2}Huazhong University of Science and Technology \\
    \normalsize{\{jingquntang, liuhongye1998, bjiang002, guanglong.hu\}@163.com,} \\ 
    \normalsize{\{wenqingzhang, yangmingkun, xbai\}@hust.edu.cn}
}
\maketitle

\begin{abstract}
Recently, transformer-based methods have achieved promising progresses in object detection, as they can eliminate the post-processes like NMS and enrich the deep representations. However, these methods cannot well cope with scene text due to its extreme variance of scales and aspect ratios. In this paper, we present a simple yet effective transformer-based architecture for scene text detection. Different from previous approaches that learn robust deep representations of scene text in a holistic manner, our method performs scene text detection based on a few representative features, which avoids the disturbance by background and reduces the computational cost. Specifically, we first select a few representative features at all scales that are highly relevant to foreground text. Then, we adopt a transformer for modeling the relationship of the sampled features, which effectively divides them into reasonable groups. As each feature group corresponds to a text instance, its bounding box can be easily obtained without any post-processing operation. Using the basic feature pyramid network for feature extraction, our method consistently achieves state-of-the-art results on several popular datasets for scene text detection.

\end{abstract}

\section{Introduction}
\label{sec:intro}

\begin{figure}
\begin{center}
\includegraphics[width=1.0\linewidth] {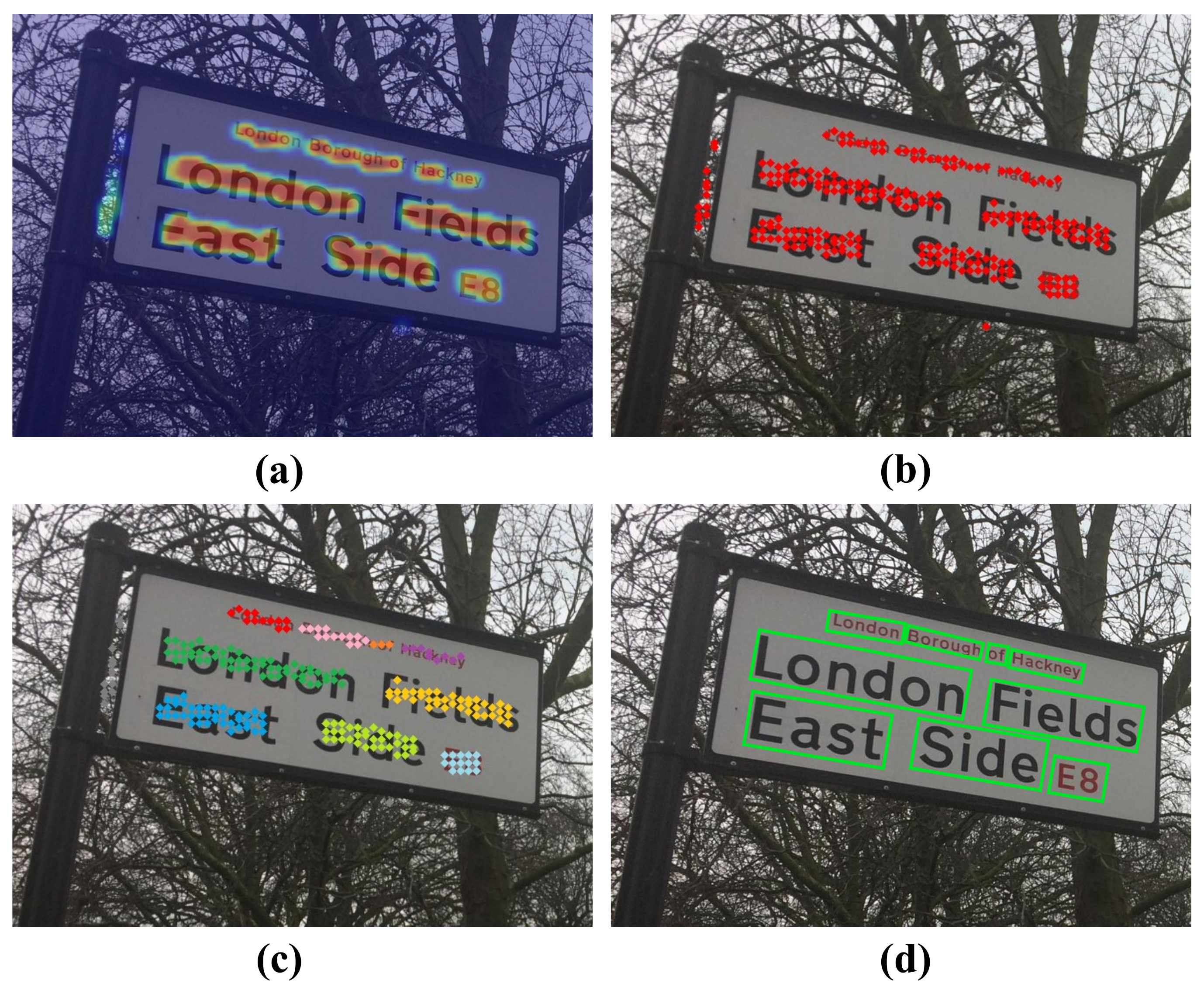}
\end{center}
  \caption{The illustration of feature sampling and grouping.
  (a) The confidence score map for text regions indicates the pixel importance for text detection.
  (b) The text features at red points containing geometric and context information of foreground text are selected by scores.
  (c) The sampled features from the same text instance are implicitly grouped at the feature level by a transformer.
  (d) The bounding boxes can be easily obtained from the grouped features.
}
\label{fig:intro}
\end{figure}

Scene text detection has been an active research field for a long time, because of its wide range of practical applications, such as scene understanding, automatic driving, and photo translation.
As a key prior component of scene text reading, scene text detection aims to precisely locate text in scene images.
Despite the noticeable improvement achieved by existing methods~\cite{MOST,RRG-Net,PAN,PSE-Net},
it is still a challenging task due to the variety of scene text, \emph{e.g.} different scales, complicated illumination, perspective distortion, multi-orientations, and complex shapes.
Moreover, most scene text detection methods depend on complicated processing to generate or refine the predicted results, such as anchor generation, non-maximum suppression (NMS)~\cite{nms}, binarization\cite{DB}, or contour extraction~\cite{suzuki1985topological}.

Inspired by the advantages of the transformer~\cite{transformer} in natural language processing (NLP), lots of works~\cite{carion2020detr,meng2021conditional,zhu2020deformable,liu2021swin,dynamic_DETR,TSP_DETR} introduce it into vision tasks to extract global-range features and model long-distance dependencies in images, while showing promising performance.
Especially in object detection, DETR-based methods~\cite{carion2020detr, zhu2020deformable, meng2021conditional} successfully use transformers to remove the complicated hand-designed processes (\emph{e.g.} NMS and anchor generation) from the former object detection frameworks~\cite{faster-rcnn, mask-rcnn,SSD}.

Although transformers bring advantages in global-range feature modeling to DETR-based frameworks~\cite{carion2020detr}, they may suffer from handling the small objects and the high computational complexity. For instance, a recent DETR-based scene text detector~\cite{raisi2021transformer} cannot achieve the satisfactory detection accuracy on the ICDAR2015 dataset~\cite{ic15} and ICDAR2017-MLT dataset~\cite{MLT17}, since the text instances in these two datasets have much larger variance of scales and aspect ratios. It is often insufficient for transformers to capture small text on the feature map at small scales, while the time cost of a DETR-based method with multi-scale feature maps is unpredictable. Essentially, unexpected background noise in higher-resolution feature maps would significantly increase the computational cost and disturb the transformer modeling.
Though, some recent works ~\cite{zhu2020deformable, meng2021conditional} improve the efficiency of transformer-based object detectors by optimizing the attention operations, they fail to achieve the competitive results in scene text detection (refer to the results reported in Tab.~\ref{tab:transformer}).

In this paper, we propose a simple yet effective transformer-based architecture for scene text detection. We argue that feature learning with the relationship of all pixels is not necessary, as foreground text instances only occupy a few small and narrow regions in scene images.
Intuitively, we firstly sample and collect the features that are highly relevant to scene text as illustrated in Fig.~\ref{fig:intro}(a)(b). Then, we adopt a transformer for modeling the relationship of the sampled features so that they can be properly grouped. As shown in Fig.~\ref{fig:intro}(c)(d), benefiting from the powerful attention mechanism of the transformer, each feature group will correspond to a text instance, which is quite convenient for predicting its bounding box.

Different from the previous scene text detection methods~\cite{EAST,DB,RRG-Net,PSE-Net,CRAFT,textboxes++} that usually learn the deep representations of scene text images in a holistic manner with CNNs, our detection method based on only a few representative features has three prominent advantages:
1) it can significantly eliminate the redundant background information, which is beneficial for improving the effectiveness and efficiency of the detection process;
2) Using a transformer to group the sampled features, we can obtain more accurate grouping results and bounding boxes without any post-processing operation;
3) As the feature sampling and grouping are implemented in an end-to-end fashion, the two stages can jointly improve the final detection performance.
To verify the effectiveness of the proposed feature sampling-and-grouping scheme, we conduct extensive experiments on several popular datasets~\cite{ic15, TD500, MLT17, MTWI, totaltext, ctw1500} for scene text detection, consistently achieving the state-of-of-art results. In addition, the comparison with the recent transformer-based detectors~\cite{carion2020detr,meng2021conditional,zhu2020deformable,raisi2021transformer} also proves the effectiveness of our method.


\begin{figure*}
\begin{center}
\includegraphics[width=1.0\linewidth] {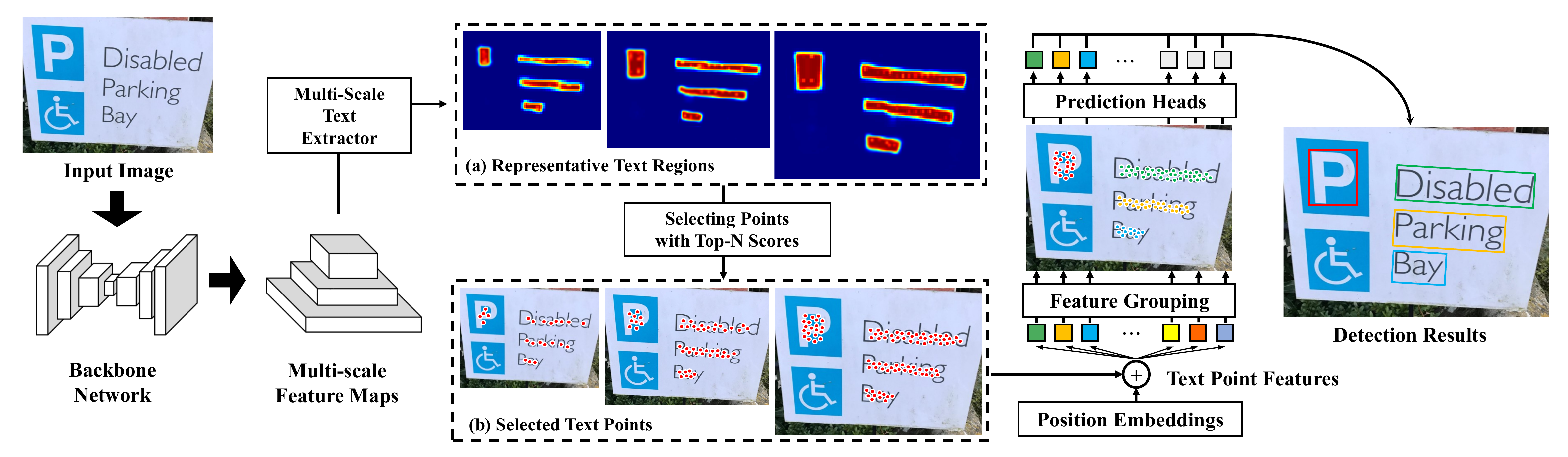}
\end{center}
\caption{The overview of our proposed transformer-based architecture.
It consists of a backbone network, a multi-scale feature sampling network, and a feature grouping network.
Specifically, multi-scale feature maps are first produced from the backbone network.
Next, a multi-scale text extractor is used to predict the confidence scores of the representative text regions at the pixel level.
Then, we select text point features with top-N scores and concatenate them with position embeddings.
After that, we adopt a transformer to model the relationship between the sampled features and implicitly group them into fine representations by the attention mechanism.
Finally, the detection results are obtained from the prediction heads.
}
\label{fig-pipeline}
\end{figure*}

\section{Related Work}
\label{sec:related}

Lots of works on scene text detection have been proposed before, which can be roughly divided into two categories: bottom-up methods and top-down methods.

\textbf{Bottom-up methods} firstly detect/segment the basic components or pixels of scene text, which are then formed into bounding boxes with some heuristic operations.
In an early method, CTPN~\cite{ctpn} develops a vertical anchor mechanism to predict sequential proposals, and naturally connects them into bounding boxes by a recurrent neural network.
To better detect long and dense text, SegLink~\cite{seglink, seglink++} detects components and links of each text instance, and combines them together to generate the final detection results.
In addition, the fundamental components can be defined as characters with affinity boxes (\emph{e.g.} CRAFT~\cite{CRAFT}) or center points with radius (\emph{e.g.} TextSnake~\cite{TextSnake}).
These methods are more flexible in detecting text with various shapes, as long as the components can be detected and grouped into final results.
However, it suffers from missing components and background noise, and the final detection results are susceptible to the grouping post-process.
Our proposed method, which is also a bottom-up method, can predict the bounding boxes by sampling and grouping at the feature level while not relying on any post-processing.

\textbf{Top-down methods} directly predict bounding boxes of scene text at the word or line level.
Inspired by the popular object detectors~\cite{SSD,faster-rcnn}, some methods~\cite{TextBoxes,textboxes++,RRPN} adjust default anchors into quadrilaterals or rotated bounding boxes to fit the multi-orientations and various aspect ratios of scene text.
EAST~\cite{EAST} directly regresses the coordinates of multi-oriented bounding boxes on the entire feature map.
To directly detect curved text in the wild, recent methods~\cite{abcnet,fourier} adopt Bezier curves or Fourier signatures for locating scene text, and apply extra processes (\emph{e.g.} BezierAlign, Inverse Fourier Transformation, and NMS) to generate the final detection results.
These top-down methods are usually more straightforward than the bottom-up ones, but they still need some hand-designed processes, such as anchor generation, NMS, and binarization.

Inspired by the power of transformers in natural language processing, the pioneer work, DETR~\cite{carion2020detr}, presents a novel transformer-based architecture for object detection.
It discards several hand-designed processes employed in~\cite{faster-rcnn, SSD, mask-rcnn}, while achieving promising performance.
Although a recent method~\cite{raisi2021transformer} has tried to apply the DETR-based architecture to scene text detection, it can not achieve a satisfying detection performance on ICDAR2015~\cite{ic15} and ICDAR2017-MLT~\cite{MLT17}.
Since scene text is more challenging than common objects for its extreme variance of scales and aspect ratios, transformers cannot obtain sufficient information from a single scale feature map.
The multi-scale scheme can somewhat cope with this problem, but it incurs a huge computational overhead for transformers.
Different from those DETR-based methods~\cite{zhu2020deformable,meng2021conditional} focusing on improving the attention units, we propose to eliminate redundant background information directly and select a few important features~\cite{zhou2002ensembling} from multi-scale feature maps.
Thus, both computational overhead and the quality of sampled features can be taken into account, which facilitates transformers being better employed for text detection.

\section{Methodology}
\label{sec:method}
In this section, we first introduce the overall architecture of the proposed scene text detection method.
Then, we elaborate on the proposed feature sampling-and-grouping scheme and further analyze the advantages of feature sampling in transformer modeling.
Finally, we describe the details about the training of our proposed method.


\subsection{Network Architecture}

As shown in Fig.~\ref{fig-pipeline}, our proposed transformer-based architecture is composed of a backbone network, a feature sampling network, and a feature grouping network.

The backbone is the basic feature pyramid network (FPN)~\cite{FPN} equipped with ResNet-50~\cite{ResNet}.
The produced feature maps $\mathbf{F}$ in three different scales (\emph{i.e.} $1/4$, $1/8$, $1/16$) are used for feature sampling.

In our feature sampling network, the three feature maps are first down-sampled to smaller scales (\emph{i.e.} $1/8$, $1/16$, $1/32$) by a Coord-Convolution layer~\cite{Coord} and a constrained deformable pooling layer.
Then, several convolution layers are employed to generate confidence score maps to distinguish representative text regions.
After that, we only select the features with top-$N_k$ scores in each scale layer $k$, and gather them into a sequence form with a shape $(\sum_{k}{N_k}, C)$, where $C$ is the channel number.

In our feature grouping network, the sampled features are first concatenated with position embeddings.
Then, we adopt transformer encoder layers to model their relationships, and implicitly aggregate the features from the same text instance.
Finally, scores and coordinates of bounding boxes (or polygons) are obtained via a text/non-text classification head and a text detection head, respectively.

\subsection{Feature Sampling}
Despite the novel structure and promising performance in object detection, transformer-based methods~\cite{carion2020detr,raisi2021transformer} can not perform well on scene text detection due to the extreme variance of scales and aspect ratios.
Following previous text detectors~\cite{DB, PSE-Net, FPN, masktextspotterv3}, we use multi-scale features from the FPN to boost the detection performance.
Nevertheless, such a scheme incurs unbearable computational cost and much longer convergence time for transformers. We observe that foreground text instances only occupy small and narrow regions, and useful information for localizing text is relatively sparse.
Hence, we propose a feature sampling network to decrease redundant background noise involved by multi-scale features, reducing the computational complexity and facilitating feature learning for transformers.

\textbf{Multi-Scale Text Extractor}
To sample representative features from foreground text, we apply a simple multi-scale text extractor to predict the confidence scores for text regions at the pixel level.
Following CoordConv~\cite{Coord}, we first concatenate each feature map with two extra channels of normalized coordinates to introduce location information.
Let $\mathbf{F}$ denote the feature maps from the FPN in different scales (\emph{i.e.} $1/4$, $1/8$, $1/16$), and
\begin{equation}
\mathbf{F} = \{ f_k \in \mathbb{R}^{H_k \times W_k \times C} | k = 0, 1, 2 \}.
\end{equation}
Then the position information is injected via
\begin{equation}
\widehat{f}_k = Conv( f_k \oplus C_k ),
\end{equation}
where $\oplus$ stands for the concatenation operation, and $C_k \in \mathbb{R}^{H_k \times W_k \times 2}$ denotes the normalized coordinates.

Inspired by deformable ROI pooling~\cite{DCN}, 
we specifically design a constrained one to down-sample the multi-scale feature maps.
Since the text area is relatively concentrated, the predicted offsets in deformable pooling with further distance will introduce irrelevant information into the pooled features.
Thus, we add a learnable scaling parameter to constrain the predicted offsets, and pool $\widehat{f}_k$ to $\Tilde{f}_k$ with smaller scales (\emph{i.e.} $1/8$, $1/16$, $1/32$).

Finally, we construct a simple scoring net $\mathbb{S}$ composed of convolution layers and a Sigmoid function to generate the confidence score maps for representative text regions at all scales.
To better distinguish the importance of pixels at different positions in each text instance, different scores over positions are used for supervision.
To generate the score maps, we adjust the Gaussian heatmap generation in general object detection~\cite{duan2019centernet,law2018cornernet} for text instances in the word level.
Specifically, a two-dimensional Gaussian distribution is implemented to generate the ground truth $\mathbf{S^{t}}=\{S_k^t | k = 0,1,2\} $ for $\mathbb{S}$, ensuring that the central part of each text instance has the highest importance score, and the scores gradually decrease from the center to contours.

\textbf{Feature Sampling}
To reduce the redundant background noise, we design a strategy for selecting representative features that are highly relevant to foreground text. These features, containing rich geometric and context information of foreground text, would be sufficient for text localization.

Let $\mathbf{S}$ denote the predicted score maps, and
\begin{equation}
\mathbf{S} = \{ S_k \in \mathbb{R}^{H_k' \times W_k'} | S_k = \mathbb{S}(\Tilde{f}_k) , k = 0, 1, 2 \}.
\end{equation}
Then, we sort scores in $S_k$, and select features with top-$N_k$ scores in $\Tilde{f}_k$ of each scale, respectively.
The selected features are gathered into $\bar{\mathbf{F}} \in \mathbb{R}^{N \times C}$ for the incoming transformer modeling:
\begin{equation}
	\bar{\mathbf{F}} = [ \bar{f}_n \in \mathbb{R}^C | n = 0, 1, ..., N],
\end{equation}
where $N = \sum\limits_{k=0}^{2}{N_k}$, and $N_k$ is the number of selected features in different scales.

Thus, the number of enormous features at all scales can be significantly reduced.
The primary selected features are probably from foreground text regions, which would contain sufficient geometric and context information for text detection.

\subsection{Feature Grouping}
Through feature selection, only a few representative features that are highly relevant to foreground text are concatenated for the incoming transformer modeling.
To reserve the position information of the sampled features,
we add the position embeddings into $\bar{\mathbf{F}}$.
Then, we adopt a transformer structure to implicitly aggregate features from the same text instance by attention mechanism.
The basic form is a stacked network with four transformer encoder layers, which are composed of self-attention modules, feed-forward layers, and layer normalization.
Following~\cite{transformer}, we construct our self-attention module as
\begin{equation}
Attn(\widehat{\mathbf{F}}) = softmax(\frac{Q(\widehat{\mathbf{F}}){K(\widehat{\mathbf{F}})}^T}{\sqrt{C'}})V(\widehat{\mathbf{F}}),
\end{equation}
where $\widehat{\mathbf{F}} \in \mathbb{R}^{N \times C'}$ denotes the sampled features with position embeddings, and $C'$ is the channel number. $Q$, $K$ and $V$ denote the different linear layers.

For previous methods~\cite{carion2020detr,meng2021conditional}, the core issue of applying the attention operation on a feature map $\boldsymbol{x} \in \mathbb{R}^{H\times W\times C'}$ is the computational complexity on all spatial locations.
In the original DETR~\cite{carion2020detr} encoder, the complexity of attention operation is $O((HW)^2C')$, which is quadratic with the spatial size.
However, in our method, it is only related to the number $N$ of selected features $\widehat{\mathbf{F}}$, and the complexity becomes $O(N^2C')$. 
In our implementation, the selected number $N^2 \ll (HW)^2$, and thus the complexity of our transformer could be significantly reduced.



Finally, the output text features are fed into two prediction heads for classification and text detection.
The text detection head is composed of fully-connected layers and a Sigmoid function.
It can regress the coordinates of rotated bounding boxes in the form of $\mathcal{B}(x,y,h,w,\theta)$ or 8 control points of Bezier-Curve~\cite{abcnet} for arbitrary-shaped text.
$x$, $y$, $h$, $w$, and $\theta$ are the coordinates of the center point, height, width, and angle, respectively.


\begin{figure*}[tb]
\centering
\includegraphics[width=1.0\linewidth]{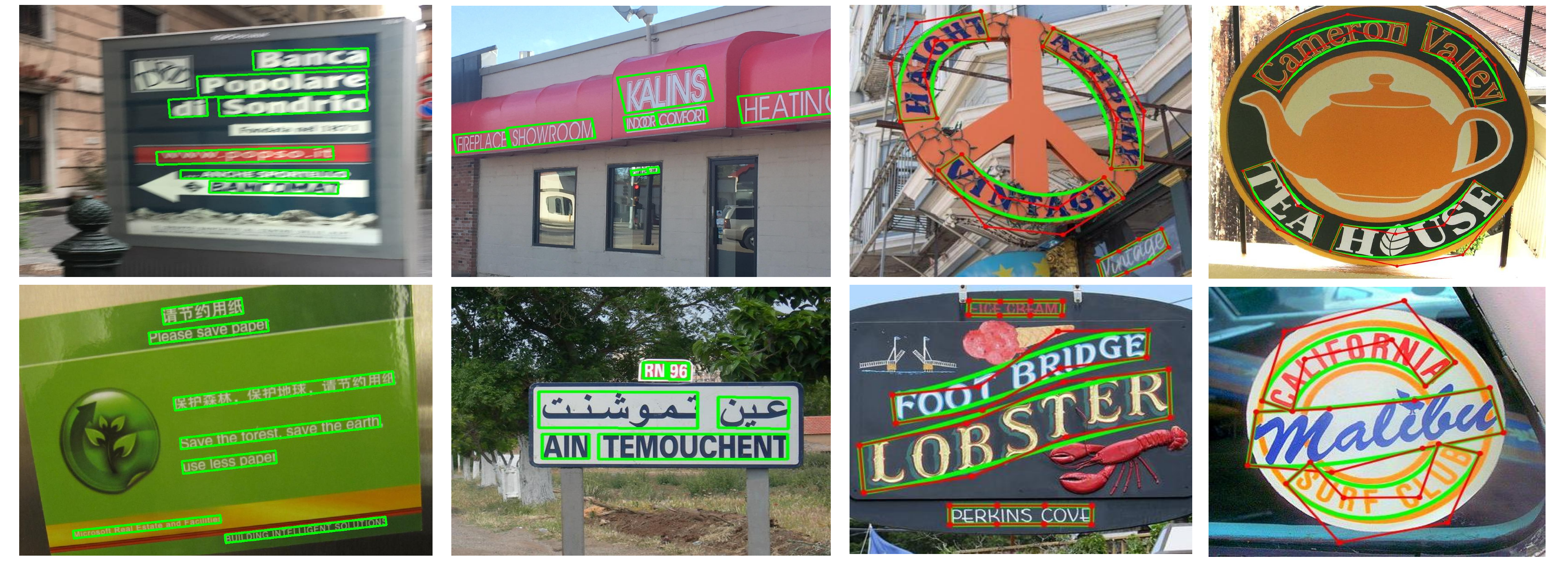}
\caption{The qualitative results of our proposed method in different cases, including multi-oriented text, long text, multi-lingual text, low-resolution text, curved text, dense text. For curved text detection, the Bezier curves' control points are drawn in red.
}
\label{fig:results}
\end{figure*}

\subsection{Optimization}
The proposed model is trained in an end-to-end manner, and the objective function consists of three parts as follows:
\begin{equation}
	\mathcal L = \lambda_c \widehat{\mathcal L}_{class} + \lambda_d \widehat{\mathcal L}_{det} + \lambda_f \mathcal L_{fs},
	\label{total_loss}
\end{equation}
where 
$\widehat{\mathcal L}_{class}$ is the loss for classification, $\widehat{\mathcal L}_{det}$ is the loss for text detection, and $\mathcal L_{fs}$ is the loss for feature selection.
$\lambda_c$, $\lambda_d$, and $\lambda_f$ are scaling factors.
Following DETR~\cite{carion2020detr}, we adopt Hungarian algorithm for pair-wise matching before calculating losses for $\widehat{\mathcal L}_{class}$ and $\widehat{\mathcal L}_{det}$.

\textbf{Loss for classification}
We adopt the Cross Entropy loss for text/non-text classification after pair-wise matching by Hungarian algorithm. It can be formulated as
\begin{equation}
	\widehat{\mathcal L}_{class} = \frac{1}{N} \sum_x-[\widehat{g}_x \cdot 	\log(\widehat{p}_x) + (1-\widehat{g}_x) \cdot \log(1-\widehat{p}_x)],
	\label{class_loss_pair}
\end{equation}
where $N$ is the total number of selected features, $\widehat{g}_x$ represents the label of sample $x$ and $\widehat{p}_x$ represents the predicted probability. The elements with $\widehat{\quad}$ denote the probabilities or labels of the matched samples after pair-wise matching.

\textbf{Loss for text detection}
For multi-oriented text detection, we adapt the Gaussian Wasserstein Distance (GWD) loss~\cite{GWD} into a scale-invariant form to better balance the loss weights of text with different scales. Due to the extreme variance of scales, the loss of small text has a negligible influence on the gradient back-propagation compared with the loss of large text.
Hence, we adjust the GWD loss as follows:
\begin{equation}
	\widehat{\mathcal L}_{det} = \frac{1}{N_r} \sum_x (1- \frac{1}{\tau + f(d^2({\frac{\widehat u_x}{|\widehat t_x|}}, {\frac{\widehat t_x}{|\widehat t_x|}}))}),
\end{equation}
where $\widehat u_x$ denotes the predicted rotated bounding box, $\widehat t_x$ denotes the target one, and $|*|$ denotes its area.
$N_r$ is the number of bounding boxes after pair-wise matching.
The elements with $\widehat{\quad}$ denote the matched bounding boxes or the target ones after pair-wise matching.
$f(\cdot)$ represents a non-linear function, and $\tau$ is a hyper-parameter to modulate the loss. $d^2$ will be explained in the Appendix.
According to the GWD loss~\cite{GWD}, we set $f(d^2)=\log{(d^2 + 1)}$ and $\tau=3$.
By normalizing $\widehat u_x$ and $\widehat t_x$ with the area of $\widehat t_x$, we can decrease the negative effect of the scale imbalance.

For arbitrary-shaped text detection, we adopt the losses for Bezier-Curve in ABC-Net~\cite{abcnet}. Thus, the prediction head for text detection is changed to two heads for predicting both bounding boxes and the control points of Bezier curves, respectively. In the bounding box prediction head, the center point coordinates, box width and box height are predicted for each bounding box $\bar{\mathcal{B}}(x,y,h,w)$. In the Bezier curve prediction head, it predicts the coordinates of 8 control points for each text instance.

\textbf{Loss for feature selection}
We apply a smooth L1 loss for optimizing the importance score maps in our feature selection as follows:
\begin{equation}
	\mathcal L_{fs} = \frac{1}{N_f} \sum_k L1_{smooth}\{S_k, S^t_k\}, k = 0, 1, 2,
\end{equation}
where $N_f$ is the total size of all score maps.
$S_k$ and $S^t_k$ are the predicted score map and the target map, respectively.

\begin{table*}[tb]
    \centering
    \setlength{\tabcolsep}{8pt}
    \begin{tabular}{lcccccccccccc}
    \toprule
    \multirow{2}{*}{Method} & \multicolumn{3}{c}{ICDAR 2015} & \multicolumn{3}{c}{MSRA-TD500} & \multicolumn{3}{c}{Total-Text} & \multicolumn{3}{c}{CTW1500}\\
    ~ & P & R & F & P & R & F& P & R & F& P & R & F \\ \midrule
    TextSnake~\cite{TextSnake} & 84.9  & 80.4  & 82.6 & 83.2  & 73.9  & 78.3 & 82.7  & 74.5  & 78.4  & 67.9  & 85.3  & 75.6 \\
    TextField~\cite{TextField} & 84.3 & 83.9 & 84.1 & 87.4 & 75.9 & 81.3  & 81.2          & 79.9          & 80.6  & 83.0          & 79.8          & 81.4  \\
    PSE-Net~\cite{PSE-Net} & 86.9 & 84.5 & 85.7  &-&-&-& 84.0          & 78.0          & 80.9   & 84.8          & 79.7          & 82.2   \\
    LOMO~\cite{LOMO} & 91.3  & 83.5  & 87.2 &-&-&-& 88.6 & 75.7  & 81.6  & 89.2 & 69.6 & 78.4\\
    CRAFT~\cite{CRAFT} & 89.8 & 84.3 & 86.9 & 88.2 & 78.2 & 82.9    & 87.6          & 79.9          & 83.6  & 86.0          & 81.1          & 83.5   \\
    PAN~\cite{PAN} & 84.0 & 81.9 & 82.9 & 84.4 & 83.8 & 84.1   & 89.3          & 81.0          & 85.0 & 86.4          & 81.2          & 83.7  \\
    DB~\cite{DB} & 91.8 & 83.2 & 87.3  & 91.5 & 79.2 & 84.9  & 87.1          & 82.5          & 84.7   & 86.9          & 80.2          & 83.4   \\
    ContourNet~\cite{ContourNet} & 87.6 & 86.1 & 86.9 &-&-&-& 86.9          & 83.9          & 85.4   & 84.1          & 83.7          & 83.9  \\
    DRRG~\cite{RRG-Net} & 88.5 & 84.7 & 86.6  & 88.1 & 82.3 & 85.1 & 86.5          & 84.9            & 85.7  & 85.9          & 83.0            & 84.5  \\
    MOST~\cite{MOST} & 89.1 & 87.3 & 88.2 & 90.4 & 82.7 & 86.4 &-&-&-&-&-&-\\
    Raisi~\emph{et al.}~\cite{raisi2021transformer} & 89.8 & 78.3 & 83.7 & 90.9 & 83.8 & 87.2 &-&-&-&-&-&- \\
    TextBPN~\cite{TextBPN} &-&-&-&86.6&84.5&85.6&90.7&85.2&87.9&86.5&83.6&85.0 \\
    \midrule
    \textbf{Ours (RBox)} & 90.9 & 87.3 & \textbf{89.1} & 91.6 & 84.8 & \textbf{88.1} &-&-&-&-&-&- \\ 
    \textbf{Ours (Bezier)} &91.1	&86.7	&88.8 &91.4	&84.7	&87.9 & 90.7 & 85.7 &\textbf{88.1}& 88.1  & 82.4	& \textbf{85.2} \\\bottomrule

    \end{tabular}
    \caption{Detection results on ICDAR2015, MSRA-TD500, Total-Text, and CTW1500. ``P", ``R", and ``F" represent Precision, Recall, and F-
measure, respectively.}
    \label{tab:benchmarks}
\end{table*}

\section{Experiments}
\label{sec:exp}

In this section, we first introduce the datasets and implementation details in our experiments.
Then, we present the evaluation results on public benchmarks and an ablation study on feature sampling.
Finally, we compare our proposed method with some popular transformer-based detection methods.

\subsection{Datasets}

\textbf{SynthText}~\cite{SynthText} is a large synthetic dataset including 800k images. It is only used to pre-train our models.

\textbf{ICDAR 2015 (IC15)}~\cite{ic15} contains 1000 training images and 500 testing images in English, most of which are severely distorted or blurred.
All images are annotated with quadrilateral boxes at the word level.

\textbf{MLT-2017 (MLT17)}~\cite{MLT17} is proposed for multi-lingual scene text detection.
It contains 7200 training images, 1800 validation images, and 9000 testing images. 
All images are annotated with quadrilateral boxes at the word level.

\textbf{MSRA-TD500}~\cite{TD500} is a multi-lingual text dataset in Chinese and English. 
It includes 300 training images and 200 testing images with multi-oriented long text.
Following previous works~\cite{MOST,TextSnake,DB}, we include HUST-TR400~\cite{TR400} as the extra training data in the fine-tuning stage.

\textbf{MTWI}~\cite{MTWI} is a large-scale dataset for Chinese and English web text reading. It contains some challenging cases, such as complex layout, small text, and watermarks.
There are 10000 training images and 10000 images for testing, and all text instances are annotated at the line level.

\textbf{Total-Text}~\cite{totaltext} is a dataset that contains text of various shapes, including horizontal, multi-oriented, and curved. It contains 1255 training images and 300 testing images, and the text instances are labeled at the word level.

\textbf{CTW1500}~\cite{ctw1500} is a curved text dataset, which consists of 1000 training images and 500 testing images. The text instances are annotated at the text-line level.

\subsection{Implementation Details}
Our model for oriented text detection is denoted as~\textbf{Ours (RBox)}, and that for arbitrary-shaped text detection is denoted as~\textbf{Ours (Bezier)}.
\textbf{Ours (RBox)} is first pre-trained on SynthText for 150 epochs, and then fine-tuned on each corresponding real-world dataset for another 100 epochs. 
\textbf{Ours (Bezier)} follows the experiment settings of ABC-Net~\cite{abcnet}, and adds its Bezier Curve Synthetic Dataset for pretraining.
We optimize our models by AdamW~\cite{adamw} with a weight decay of $1e^{-4}$ and a momentum of 0.9.
The initial learning rate for pre-training and fine-tuning is $1e^{-3}$ and $5e^{-4}$, respectively. Both of them will decay to $1e^{-4}$ after the 40th epoch.
More details can be referred to Appendix.

\subsection{Evaluation on Benchmarks}
To compare with previous scene text detectors, we evaluate our proposed method on several popular benchmarks for scene text detection.
We adopt the best model configuration in the \#5 of Tab.~\ref{tab:ablation} for evaluating on all benchmarks.
As shown in Fig.~\ref{fig:results}, we provide some qualitative results in different cases, including multi-oriented text, long text, multi-lingual text, small text, low-resolution text, and curved text.

\noindent \textbf{Multi-oriented text detection}
We evaluate our method for multi-oriented text on the IC15 dataset and the MSRA-TD500 dataset, which contain lots of small, low-resolution, and long text instances.
As shown in Tab.~\ref{tab:benchmarks}, our model outperforms previous state-of-the-art method by 0.9\% on both IC15 and MSRA-TD500.
Compared with the former DETR-based method~\cite{raisi2021transformer}, our proposed model shows a much better detection performance (89.1\% \emph{vs.} 83.7\%) on small and blurry text of IC15.
Compared with previous CNN-based methods on MSRA-TD500, our method outperforms them by at least 1.7\% in terms of f-measure, owing to the advantages of transformers in extracting global-range features and long-distance dependencies.

\noindent \textbf{Curved text detection}
To prove our method's effectiveness on curved text, we evaluate it on two popular curved text benchmarks, \emph{i.e.} the Total-Text dataset and the CTW1500 dataset. 
As shown in Tab.~\ref{tab:benchmarks}, our method obtains 0.2\% improvement in terms of f-measure compared with the state-of-the-art method TextBPN~\cite{TextBPN}.
With the help of Bezier-Curve~\cite{abcnet}, our method could generate polygons for curved text, which can not be precisely detected by the former DETR-based method~\cite{raisi2021transformer}.
Moreover, our method with Bezier-Curve could also achieve state-of-the-performance performance on the IC15 and MSRA-TD500 datasets.

\begin{table}[tb]
\centering
\begin{tabularx}{1.0\linewidth}{@{}l*{4}X@{}}
\toprule
Method & P & R & F & FPS  \\ \midrule
Corner~\cite{Corner} & 83.8 & 55.6 & 66.8 & - \\
CRAFT~\cite{CRAFT} & 80.6 & 68.2 & 73.9 & 8.6 \\
PSE-Net~\cite{PSE-Net} & 73.8 & 68.2 & 70.7 & - \\
DB~\cite{DB} & 83.1 & 67.9 & 74.7 & \textbf{19.0} \\
DRRG~\cite{RRG-Net} & 75.0 & 61.0 & 67.3 & -  \\
Xiao~\emph{et al.}~\cite{xiao} & 84.2 & 72.8 & 78.1 & - \\
MOST~\cite{MOST} & 82.0 & 72.0 & 76.7 & 10.1  \\
Raisi~\emph{et al.}~\cite{raisi2021transformer} & 84.8 & 63.2 & 72.4 & - \\
\midrule
\textbf{Ours (RBox)} & 87.3 & 73.2 & \textbf{79.6} & 13.1 \\ \bottomrule
\end{tabularx}
\caption{Detection results on the MLT-2017 test dataset.}
\label{tab:MLT17}
\end{table}

\noindent \textbf{Multi-lingual text detection}
To demonstrate the robustness of our model for different languages, we evaluate it on two large-scale scene text datasets (\emph{i.e.} the MLT17 test dataset and the MTWI dataset).
As shown in Tab.~\ref{tab:MLT17}, 
compared with the state-of-the-art model~\cite{xiao}, our model obtains 3.1\%, 0.4\%, and 1.5\% improvements in terms of precision, recall, and f-measure, respectively.
We also evaluate our model on the MTWI dataset, which contains multi-lingual text from web images.
Our method achieves the best performance 75.2\% in terms of f-measure with a competitive inference speed (21.5 FPS).

\begin{table}[tb]
\centering
\begin{tabularx}{1.0\linewidth}{@{}l*{4}X@{}}
\toprule
Method & P & R & F & FPS  \\ \midrule
SegLink *~\cite{seglink} & 70.0 & 65.4 & 67.6 & - \\
TextBoxes++ *~\cite{textboxes++} & 66.8 & 56.3 & 61.1 & - \\
Seglink++~\cite{seglink++} & 74.7 & 69.7 & 72.1 & - \\
BDN~\dag~\cite{BDN} & 77.3 & 70.0 & 73.4 & 2.7 \\
PAN~\dag~\cite{PAN} & 78.9 & 68.9 & 73.5 & 16.9 \\
MOST~\cite{MOST} & 78.8 & 71.1 & 74.7 & \textbf{23.5} \\
\midrule
\textbf{Ours (RBox)} & 78.4 & 72.3 & \textbf{75.2} & 21.5 \\ \bottomrule
\end{tabularx}
\caption{Detection results on the MTWI dataset. * and \dag indicate that the results are reported by SegLink++~\cite{seglink++} and MOST~\cite{MOST}, respectively.}
\label{tab:MTWI}
\end{table}

\subsection{Experiments on Feature Sampling}
To demonstrate the effectiveness of our proposed feature sampling scheme, we conduct several experiments with different sampling configurations on the IC15 dataset and the MLT17 validation dataset.
As shown in \#1, \#2, and \#5 of Tab.~\ref{tab:ablation}, our method can significantly improve the performance with the help of higher-resolution feature maps.
For IC15, sampling features at all scales outperforms the other two configurations by 22.1\% and 6.9\%, respectively.
Consistently, it achieves 15.6\% and 6.1\% performance gain compared with others on MLT17.
In addition, we conduct four configurations to explore the effects of sampling numbers from \#3 to \#6 in Tab.~\ref{tab:ablation}.
We observe that the performance can increase with more sampling features, but stagnates in the last.
The models with fewer sampled features can not perform well, because these features do not contain enough geometric and context information of all text instances.
From \#5 and \#6, we find the performance slightly decreases as the sampling number increases, which may introduce more redundant features and incur negative effects.

To further evaluate the impact of sampling points, we we adopt an adaptive sampling scheme for training in \#7.
For every training image, we sort all features from the foreground text area by the predicted scores, and sample a fixed percentage (25\%) of them with top scores.
In this way, the sampling number is adaptive to the foreground feature number, and the performance of adaptive sampling is close to \#5 and \#6. Hence, our method is not sensitive when the sampling number is larger than that of \#5.
Moreover, we try to use all the features in different scales for the transformer modeling, but encounter the issue of ``Out Of Memory" during training.
Assuming the size of input images is $1024 \times 1024$, the sizes of $L0$, $L1$, and $L2$ would be $32\times32$, $64\times64$, and $128\times128$, respectively.
The whole features mixed with background are difficult to model, and lead to a huge computational cost which is nearly 1400 times more than that of \#5.
Thus, our feature sampling is effective to decrease complexity for multi-scale feature maps and preserve the important information for scene text detection.


\begin{table}[tb]
	\centering
	\small
	\setlength\tabcolsep{3.5pt}
	\begin{tabularx}{1\linewidth}{l|ccc|ccc|ccc}
		\hline
		\multirow{2}*{ID} &\multicolumn{3}{c}{Sampled Features} \vline & \multicolumn{3}{c}{IC15} \vline &\multicolumn{3}{c}{MLT17 val} \\
		~&L0 & L1 & L2 & P & R & F & P & R & F\\ \hline
		\#1&64 & - & - &75.2&60.4&	67.0& 	79.9&	53.2&	63.9 \\
		\#2&64 & 128& -	&86.5&78.3&	82.2& 	82.7&	65.9&	73.4 \\
		\#3&16  & 32  & 64  & 82.4 &	73.7&	77.8&	78.9	&61.1	&68.9 \\ 
		\#4&32  & 64  & 128	& 88.1 &	84.0 &	86.0 & 84.1	& 72.8	&78.0 \\
		\#5&64  & 128 & 256	& 90.9 &	87.3 &	\textbf{89.1} & 86.8& 73.4	& \textbf{79.5} \\
		\#6&128 & 256 & 512 & 90.2  &87.9	&89.0& 85.9	&73.8	&79.4\\
 		\#7 &\multicolumn{3}{c|}{Adaptive Sampling} &90.7	&87.2&	88.9 & -	&-	&-\\
		\hline
	\end{tabularx}
	\caption{The experiments of feature sampling number on the IC15 test dataset and the MLT17 validation dataset. ``L0", ``L1" and ``L2" denote the feature maps in different scales (\emph{i.e.} 1/32, 1/16, 1/8).
	}
	\label{tab:ablation}
\end{table}

\begin{figure}[tb]
\begin{center}
\includegraphics[width=1.0\linewidth] {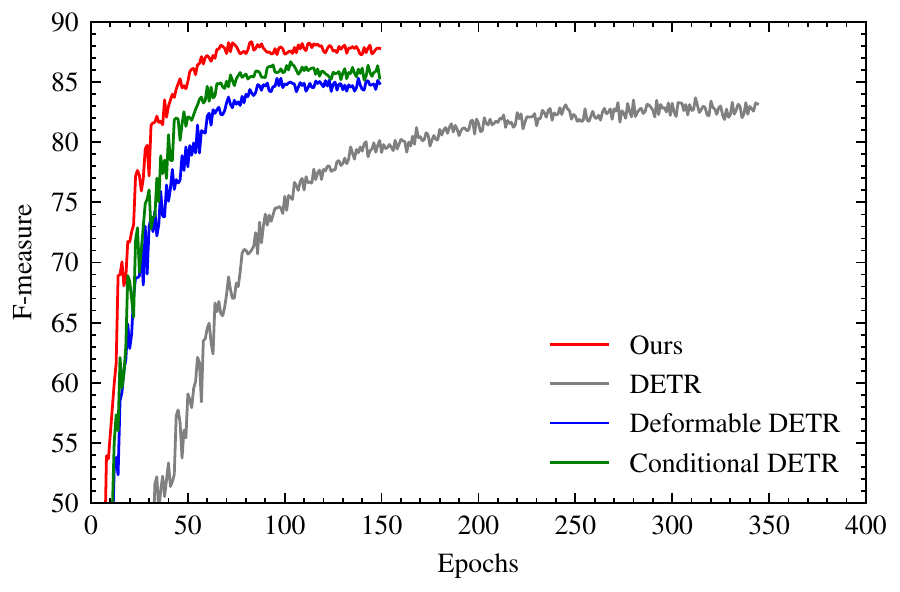}
\end{center}
  \caption{The convergence curves for DETR, Deformable DETR, Conditional DETR and Ours (RBox) on SynthText.
  The training and validation set is split from SynthText with a ratio 8:2.
  We train the previous methods by adjusting their official codes for multi-oriented text detection and follow the same settings as ours.
}
\label{fig:convergence}
\end{figure}

\subsection{Comparisons with Transformer-Based Detection Methods}
In this part, we compare our model with some popular transformer-based methods (\emph{i.e.} DETR~\cite{carion2020detr}, Deformable DETR~\cite{zhu2020deformable}, and Conditional DETR~\cite{conditionaldetr}) in object detection.
We use their official codes and follow our training settings for fair comparisons. Noticeably, we adjust their codes for multi-oriented text detection by adding angle regression and using our loss function.

Since pre-training on SynthText is a necessary step in previous methods~\cite{MOST,DB,RRG-Net,ContourNet}, we first compare the convergence speed on SynthText.
We train most models excluding DETR with the same training settings as ours, but train DETR for 350 epochs for its low convergence speed.
As illustrated in Fig.~\ref{fig:convergence}, the convergence speed of our method is much faster than DETR, because ours can significantly reduce redundant information and ease transformer modeling.
Compared with the other two methods~\cite{zhu2020deformable,conditionaldetr} focusing on increasing the efficiency of the attention units, our feature sampling-and-grouping scheme has a simpler pipeline but demonstrates a competitive convergence speed with a better detection performance.
After fine-tuning, our proposed model obtains the best detection performance in terms of f-measure on IC15 and MLT17 as shown in Tab.~\ref{tab:transformer}.

In addition, we compare the FLOPs, the number of model parameters, and the inference speed with the previous transformer-based methods.
For a fair comparison, we resize both sides of input images to $640$ for all models to calculate the FLOPs, and use the same images from the IC15 test dataset to measure the inference speed by FPS.
The number of object queries is set to 100 for previous methods, and we adopt the \#5 configuration in Tab.~\ref{tab:ablation} for ours.
As shown in Tab.~\ref{tab:flopsandparams}, our proposed transformer-based architecture has a lower computational cost in terms of FLOPs and a faster inference speed.

\begin{table}[tb]
	\centering
	\small
	\setlength\tabcolsep{4pt}
	\begin{tabularx}{1\linewidth}{c|ccc|ccc}
		\hline
		\multirow{2}*{Transformer Layer} & \multicolumn{3}{c}{IC15} \vline &\multicolumn{3}{c}{MLT17 val} \\
		~ & P & R & F & P & R & F\\ 	\hline
		Basic Layer &90.8	&87.3	&89.1 	&86.8 &73.4 &79.5  \\ 
		Swin Transformer Layer	&90.9	&88.1	&\textbf{89.5} 	&87.2	&73.4	&\textbf{79.7}   \\ 
	
		\hline
	\end{tabularx}
	\caption{The experiment on the transformer layers in our feature grouping network.
	}
	\label{tab:swin}
\end{table}

\subsection{Transformer Structure}
Despite the state-of-the-art performance achieved by our basic model architecture, we replace the basic transformer encoder layers with those in the modern transformer structure, \emph{i.e.} Swin-Transformer~\cite{liu2021swin}, for further improvement.
Different from applying Swin-Transformer for images, we only use four swin-transformer blocks for our feature grouping.
Since it is designed for 2-D feature maps, we feed the feature map into the swin-transformer stage while masking out the unsampled features.
Owing to the power of Swin-Transformer layers, our model obtains 0.4\% and 0.2\% performance gain on the IC15 and the MLT17 datasets as shown in Tab.~\ref{tab:swin}.

\begin{table}[tb]
	\centering
	\small
	\setlength\tabcolsep{4pt}
	\begin{tabularx}{1\linewidth}{c|ccc|ccc}
		\hline
		\multirow{2}*{Methods} & \multicolumn{3}{c}{IC15} \vline &\multicolumn{3}{c}{MLT17 val} \\
		~ & P & R & F & P & R & F\\ \hline
		DETR*~\cite{carion2020detr}  & 87.9&	75.4&	81.2&	84.6	&63.4&	72.5 \\ 
		Deformable DETR*~\cite{zhu2020deformable}	& 88.3 &	84.7 &	86.5 & 86.5	&69.3	&77.0 \\
		Conditional DETR*~\cite{conditionaldetr}	& 87.5 &	81.8 &	84.6 & 85.9	&67.8	&75.8 \\
		Raisi~\emph{et al.}~\cite{raisi2021transformer}	& 89.8 & 78.3 & 83.7 & - & - & - \\
		\midrule
		\textbf{Ours (RBox)} & 90.9 & 87.3 & \textbf{89.1} & 86.8& 73.4	& \textbf{79.5}\\
		\hline
	\end{tabularx}
	\caption{Comparisons with transformer-based methods on the IC15 test dataset and the MLT17 validation dataset. * indicates the methods are trained by adjusting their official codes for multi-oriented text detection.
	}
	\label{tab:transformer}
\end{table}

\begin{table}[tb]
\centering
\begin{tabularx}{1.0\linewidth}{cccc}
\toprule
Method & FLOPs & Params & FPS \\
\midrule
DETR~\cite{carion2020detr}  & 38.9G & 41.3M& 9.7 \\
Deformable DETR~\cite{zhu2020deformable} & 36.8G & 39.8M & 7.6 \\
Conditional DETR~\cite{conditionaldetr} & 42.2G & 43.2M& 9.1 \\
\midrule
\textbf{Ours (RBox)} & 35.9G & 38.3M & 12.9 \\
\bottomrule
\end{tabularx}
\caption{Comparisons with transformer-based methods on FLOPs, the number of parameters, and the inference speed.
For FLOPs, both sides of input images are set to 640.
For FPS, we evaluate all models on the IC15 test dataset with the same inference setting of ours.
The number of object queries is set to 100 for previous methods, and we adopt the \#5 configuration in Tab.~\ref{tab:ablation} for ours.
}
\label{tab:flopsandparams}
\end{table}

\begin{table*}[tb]
\centering
\footnotesize
\setlength\tabcolsep{2.2pt}
{
\begin{tabularx}{1.0\linewidth}{c|lccccccccccccccccc|c}
\hline
 & Method & Backbone & MS & PL & BD & BR & GTF & SV & LV & SH & TC & BC & ST & SBF & RA & HA & SP & HC & AP$_{50}$   \\
\hline
\multirow{8}*{\rotatebox{90}{Two-stage}} & ICN~\cite{ICN} & R-101 & $\surd$ & 81.40 & 74.30 & 47.70 & 70.30 & 64.90 & 67.80 & 70.00 & 90.80 & 79.10 & 78.20 & 53.60 & 62.90 & 67.00 & 64.20 & 50.20 & 68.20 \\
& RoI-Trans.~\cite{ROI-Trans} & R-101 & $\surd$ & 88.64 & 78.52 & 43.44 & 75.92 & 68.81 & 73.68 & 83.59 & 90.74 & 77.27 & 81.46 & 58.39 & 53.54 & 62.83 & 58.93 & 47.67 & 69.56 \\
& SCRDet~\cite{SCRDet} & R-101 & $\surd$ & 89.98 & 80.65 & 52.09 & 68.36 & 68.36 & 60.32 & 72.41 & 90.85 & \textbf{\textcolor{red}{87.94}} & 86.86 & 65.02 & 66.68 & 66.25 & 68.24 & 65.21 & 72.61 \\
& Gliding Vertex~\cite{Gliding} & R-101 &  & 89.64 & 85.00 & 52.26 & 77.34 & 73.01 & 73.14 & 86.82 & 90.74 & 79.02 & 86.81 & 59.55 & \textbf{\textcolor{blue}{70.91}} & 72.94 & 70.86 & 57.32 & 75.02 \\
& CenterMap OBB~\cite{CenterMapOBB} & R-101 & $\surd$ & 89.83 & 84.41 & 54.60 & 70.25 & 77.66 & 78.32 & 87.19 & 90.66 & 84.89 & 85.27 & 56.46 & 69.23 & 74.13 & 71.56 & 66.06 & 76.03 \\
& FPN-CSL~\cite{FPN-CSL} & R-152 & $\surd$ & 90.25 & \textbf{\textcolor{blue}{85.53}} & 54.64 & 75.31 & 70.44 & 73.51 & 77.62 & 90.84 & 86.15 & 86.69 & 69.60 & 68.04 & 73.83 & 71.10 & 68.93 & 76.17 \\
& RSDet-II~\cite{RSDet-II} & R-152 & $\surd$ & 89.93 & 84.45 & 53.77 & 74.35 & 71.52 & 78.31 & 78.12 & \textbf{\textcolor{red}{91.14}} & 87.35 & 86.93 & 65.64 & 65.17 & 75.35 & \textbf{\textcolor{red}{79.74}} & 63.31 & 76.34 \\ \cline{2-20}
& \multirow{4}{*}{Oriented R-CNN~\cite{orientedrcnn}}   & R-50 & &89.46 &82.12 &54.78 &70.86 &78.93 &83.00 &88.20 &\textbf{\textcolor{blue}{90.90}} &87.50 &84.68 &63.97 &67.69 &74.94 &68.84 &52.28 &75.87 \\
& & R-101 & &88.86 &83.48 &55.27 &76.92 &74.27 &82.10 &87.52 &\textbf{\textcolor{blue}{90.90}} &85.56 &85.33 &65.51 &66.82 &74.36 &70.15 &57.28 &76.28 \\
& & R-50 & $\surd$ & 89.84 &85.43 &61.09 &79.82 &79.71 &\textbf{\textcolor{red}{85.35}} &88.82 &90.88 &86.68 &\textbf{\textcolor{blue}{87.73}} &72.21 &70.80 &\textbf{\textcolor{blue}{82.42}} &78.18 &74.11 &\textbf{\textcolor{red}{80.87}} \\
& & R-101 & $\surd$ & \textbf{\textcolor{blue}{90.26}} &84.74 &\textbf{\textcolor{red}{62.01}} &80.42 &79.04 &\textbf{\textcolor{blue}{85.07}} &88.52 &90.85 &87.24 &\textbf{\textcolor{red}{87.96}} &72.26 &70.03 &\textbf{\textcolor{red}{82.93}} &78.46 &68.05 &80.52 \\
\hline
\multirow{8}*{\rotatebox{90}{Refine-stage}} & CFC-Net~\cite{CFCNet} & R-101 & $\surd$ & 89.08 & 80.41 & 52.41 & 70.02 & 76.28 & 78.11 & 87.21 & 90.89 & 84.47 & 85.64 & 60.51 & 61.52 & 67.82 & 68.02 & 50.09 & 73.50 \\
& DCL~\cite{DCL} & R-152 & $\surd$ & 89.26 & 83.60 & 53.54 & 72.76 & 79.04 & 82.56 & 87.31 & 90.67 & 86.59 & 86.98 & 67.49 & 66.88 & 73.29 & 70.56 & 69.99 & 77.37\\
& RIDet~\cite{RIDet} & R-50 & $\surd$ & 89.31 & 80.77 & 54.07 & 76.38 & \textbf{\textcolor{blue}{79.81}} & 81.99 & \textbf{\textcolor{blue}{89.13}} & 90.72 & 83.58 & 87.22 & 64.42 & 67.56 & 78.08 & 79.17 & 62.07 & 77.62\\
& S$^2$A-Net~\cite{S2ANet} & R-101 & $\surd$ & 89.28 & 84.11 & 56.95 & 79.21 & \textbf{\textcolor{red}{80.18}} & 82.93 & \textbf{\textcolor{red}{89.21}} & 90.86 & 84.66 & 87.61 & 71.66 & 68.23 & 78.58 & 78.20 & 65.55 & 79.15\\
& R$^3$Det-GWD~\cite{R3Det-GWD} & R-152 & $\surd$ & 89.66 & 84.99 & 59.26 & \textbf{\textcolor{red}{82.19}} & 78.97 & 84.83 & 87.70 & 90.21 & 86.54 & 86.85 & \textbf{\textcolor{blue}{73.04}} & 67.56 & 76.92 & 79.22 & 74.92 & 80.19 \\ \cline{2-20}
& \multirow{2}{*}{R$^3$Det-KLD~\cite{R3Det-KLD}} & R-50 & $\surd$ & 89.90 & 84.91 & 59.21 & 78.74 & 78.82 & 83.95 & 87.41 & 89.89 & 86.63 & 86.69 & 70.47 & 70.87 & 76.96 & \textbf{\textcolor{blue}{79.40}} & \textbf{\textcolor{blue}{78.62}} & 80.17 \\
& ~ & R-152 & $\surd$ & 89.92 & 85.13 & 59.19 & \textbf{\textcolor{blue}{81.33}} & 78.82 & 84.38 & 87.50 & 89.80 & 87.33 & 87.00 & 72.57 & \textbf{\textcolor{red}{71.35}} & 77.12 & 79.34 & \textbf{\textcolor{red}{78.68}} & \textbf{\textcolor{blue}{80.63}} \\
\hline
\multirow{6}*{\rotatebox{90}{Single-stage}} & PolarDet~\cite{PolarDet} & R-101 & $\surd$ & 89.65 & \textbf{\textcolor{red}{87.07}} & 48.14 & 70.97 & 78.53 & 80.34 & 87.45 & 90.76 & 85.63 & 86.87 & 61.64 & 70.32 & 71.92 & 73.09 & 67.15 & 76.64 \\
& RDD~\cite{RDD} & R-101 & $\surd$ & 89.15 & 83.92 & 52.51 & 73.06 & 77.81 & 79.00 & 87.08 & 90.62 & 86.72 & 87.15 & 63.96 & 70.29 & 76.98 & 75.79 & 72.15 & 77.75\\
& GWD~\cite{GWD} & R-152 & $\surd$ & 89.06 & 84.32 & 55.33 & 77.53 & 76.95 & 70.28 & 83.95 & 89.75 & 84.51 & 86.06 & \textbf{\textcolor{red}{73.47}} & 67.77 & 72.60 & 75.76 & 74.17 & 77.43\\ \cline{2-20}
& \multirow{2}{*}{KLD~\cite{KLD}} & R-50 & & 88.91 & 83.71 & 50.10 & 68.75 & 78.20 & 76.05 & 84.58 & 89.41 & 86.15 & 85.28 & 63.15 & 60.90 & 75.06 & 71.51 & 67.45 & 75.28\\
& ~ & R-50 & $\surd$ & 88.91 & 85.23 & 53.64 & 81.23 & 78.20 & 76.99 & 84.58 & 89.50 & 86.84 & 86.38 & 71.69 & 68.06 & 75.95 & 72.23 & 75.42 & 78.32\\ \cline{2-20}
& \multirow{2}{*}{\textbf{Ours (RBox)}} & R-50 &  & \textbf{\textcolor{red}{90.36}} & 85.31 & 56.39 & 76.45 & 74.55 & 83.46 & 87.78 & 90.86 & 85.85 & 85.28 & 64.52 & 67.82 & 77.72 & 74.32 & 67.80 & 77.90 \\
& ~ & R-50 & $\surd$ & 89.81 & 85.19 & \textbf{\textcolor{blue}{61.35}} & 76.18 & 79.29 & 84.81 & 88.26 & 90.86 & \textbf{\textcolor{blue}{87.55}} & 87.42 & 66.89 & 70.10 & 78.40 & 79.28 & 68.48 & 79.59 \\
\hline
\end{tabularx}}
\caption{Detection results on the DOTA-v1.0 testing set. R-50, R-101, and R-152 denote ResNet-50, ResNet-101, and ResNet-152, respectively.
MS indicates that multi-scale testing is used. \textcolor{red}{Red} and \textcolor{blue}{blue} indicate the top two performances.}
\label{tab:dota}
\end{table*}

\begin{figure*}[t]
	\begin{center}
		\includegraphics[width=1.0\linewidth] {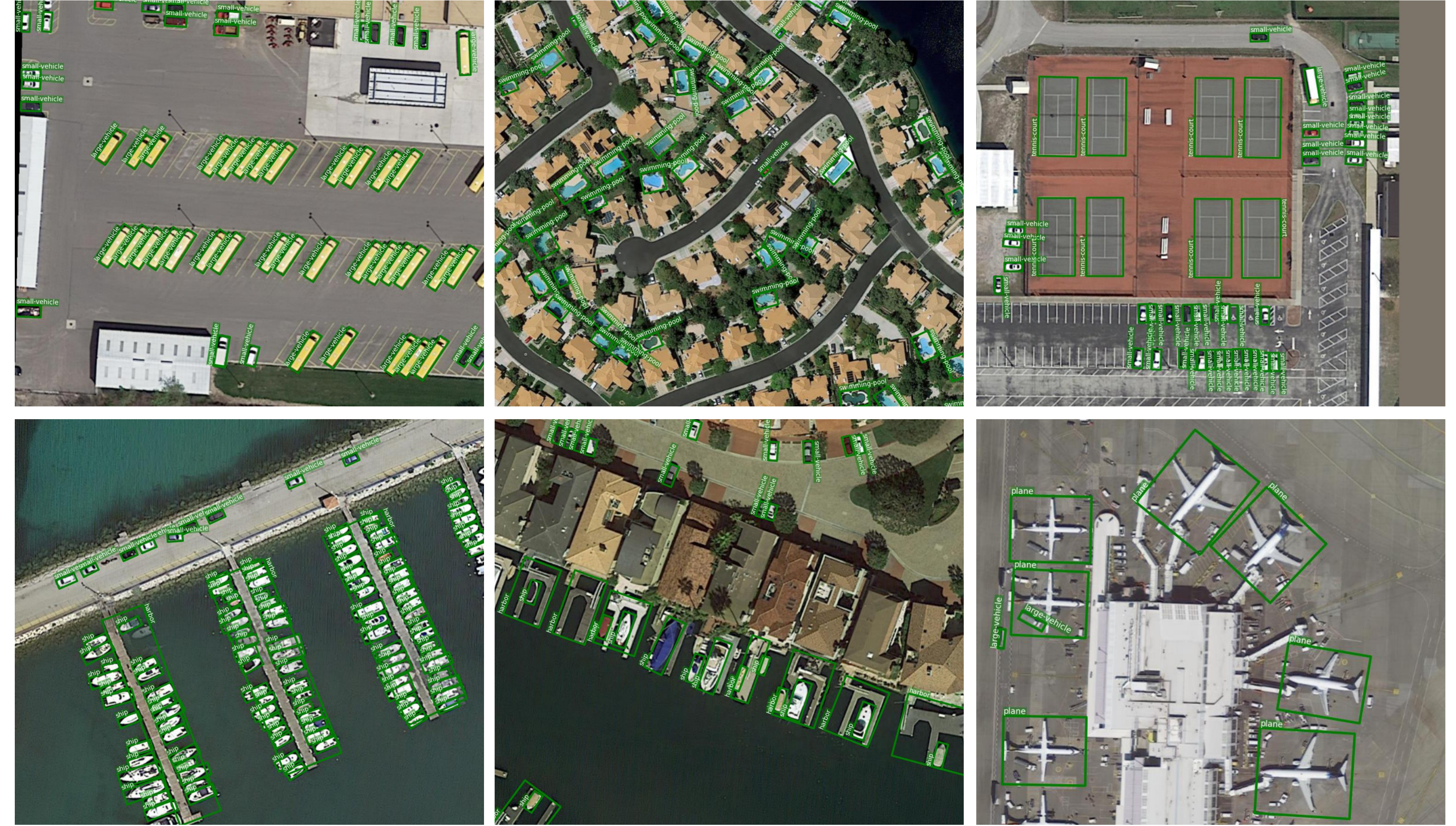}
	\end{center}
	\caption{The qualitative results on DOTA-v1.0 testing set. It contains 15 common categories, such as large-vehicle, small-vehicle, plane, swimming-pool, ship, tennis-court, etc.}
	\label{fig:dota}
\end{figure*}

\subsection{Rotated Object Detection}
Our proposed method not only achieves state-of-the-art performance on scene text detection, but also performs well on oriented object detection.
To prove the effectiveness of our method, we adapt it to oriented object detection and evaluate it on a popular dataset for oriented object detection in aerial images, \emph{i.e.,} DOTA-v1.0~\cite{DOTA}.
DOTA-v1.0 is one of the largest dataset for oriented object detection in aerial images, and it contains 15 common categories, 2806 images and 188282 instances.

In the training, we use the same loss function as the loss for multi-oriented text detection. The feature sampling scheme is consistent with the configuration \#5.
Following the pre-processing in previous methods~\cite{KLD,GWD}, we split the training images of DOTA-v1.0 into 1024 × 1024 sub-images with an overlap of 200 pixels. We train our model for 100 epochs with an initial learning rate $1e^{-4}$, and decay it at 50th and 80th epoch, respectively.

As shown in Tab.~\ref{tab:dota}, we compare our model with previous oriented object detection approaches in both single-scale and multi-scale testing manners. For a fair comparison, our method achieves the best performance among the single-stage approaches, and outperform KLD~\cite{KLD} by 1.32 AP$_{50}$.
By multi-scale testing, our model also achieves the competitive result 79.59 in terms of AP$_{50}$ with refine-stage and two-stage approaches.

\subsection{Limitation}
For our feature sampling-and-grouping scheme, it is hard to deal with the ``text overlapping" cases, which mean two text instances overlap each other.
Although our feature grouping network can model the relationship of the sampled features, the features of the overlapping text are quite complex and tangled. Thus, our proposed method sometimes fails in these cases, which are shown in the Appendix.

\section{Conclusion}
\label{sec:conclusion}
In this paper, we present a simple yet effective transformer-based architecture for scene text detection.
Different from previous methods in scene text detection, our method leverages only a few representative features containing sufficient geometric and context information of foreground text.
It is able to effectively reduce the redundant background noise and overcome the complexity limitation of the self-attention module.
With the power of transformers, we can obtain more accurate bounding boxes without any post-processing.
Through extensive experiments on several benchmarks, we demonstrate the effectiveness of our proposed method by consistently achieving state-of-the-art results on both multi-oriented text datasets and arbitrary-shaped text datasets.

\noindent \textbf{Acknowledgement} This work was supported by National
Key R\&D Program of China (No. 2018YFB1004600).

{\small
\bibliographystyle{ieee_fullname}
\bibliography{egbib}
}

\clearpage

\appendix

\section{Implementation Details}

\subsection{Network Architecture}
Our proposed transformer-based architecture is composed of a backbone network, a feature sampling network, and a feature grouping network.

\begin{figure}[tb]
	\begin{center}
		\includegraphics[width=0.9\linewidth] {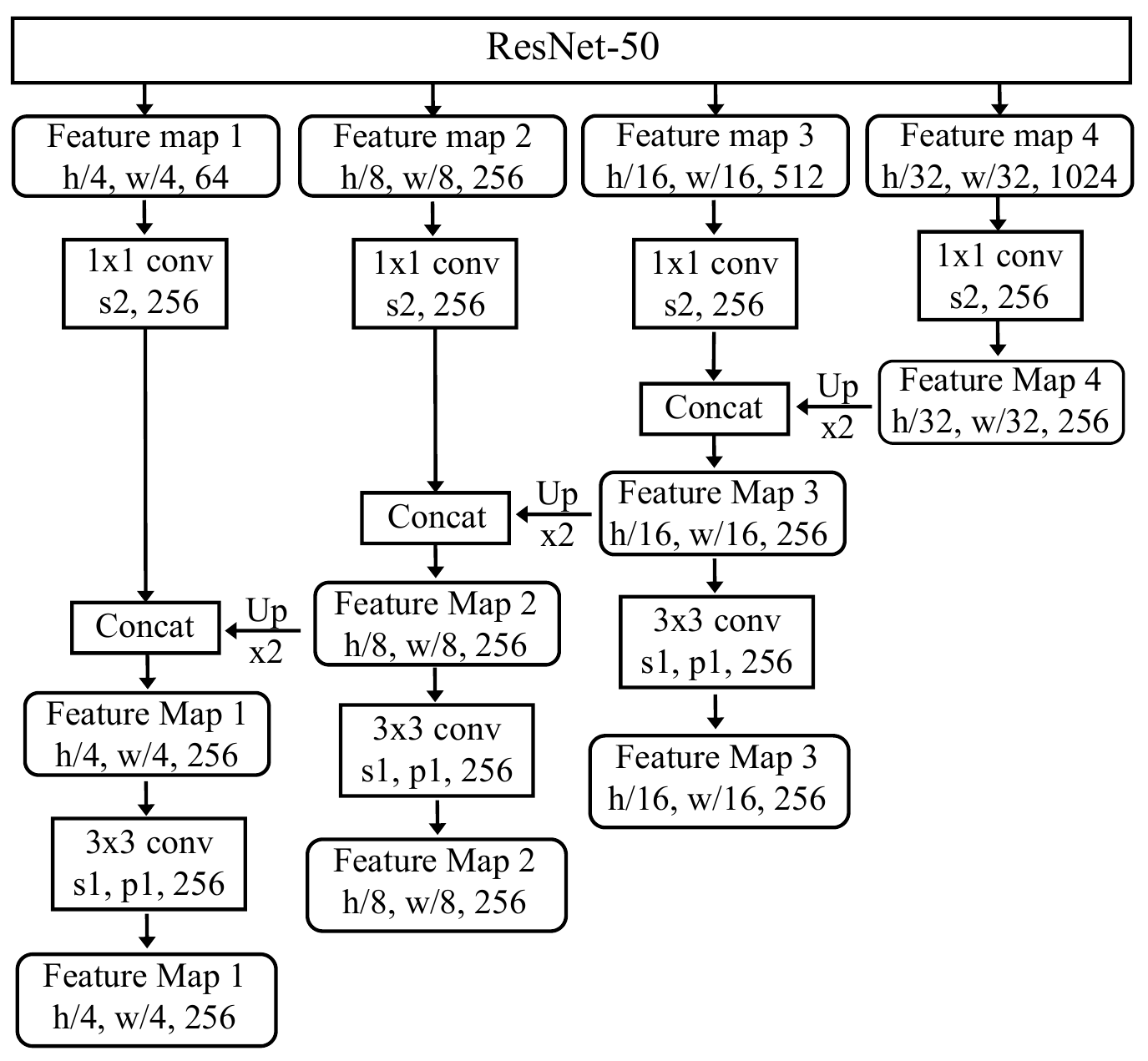}
	\end{center}
	\caption{The structure of our feature pyramid network equipped with ResNet-50.
	}
	\label{fig:FPN}
\end{figure}

The backbone is the basic feature pyramid network (FPN)~\cite{FPN} equipped with ResNet-50~\cite{ResNet} as shown in Fig.~\ref{fig:FPN}, 
The produced feature maps in three different scales (\emph{i.e.} $1/4$, $1/8$, $1/16$) are used for feature sampling.

\begin{figure}[tb]
	\begin{center}
		\includegraphics[width=0.9\linewidth] {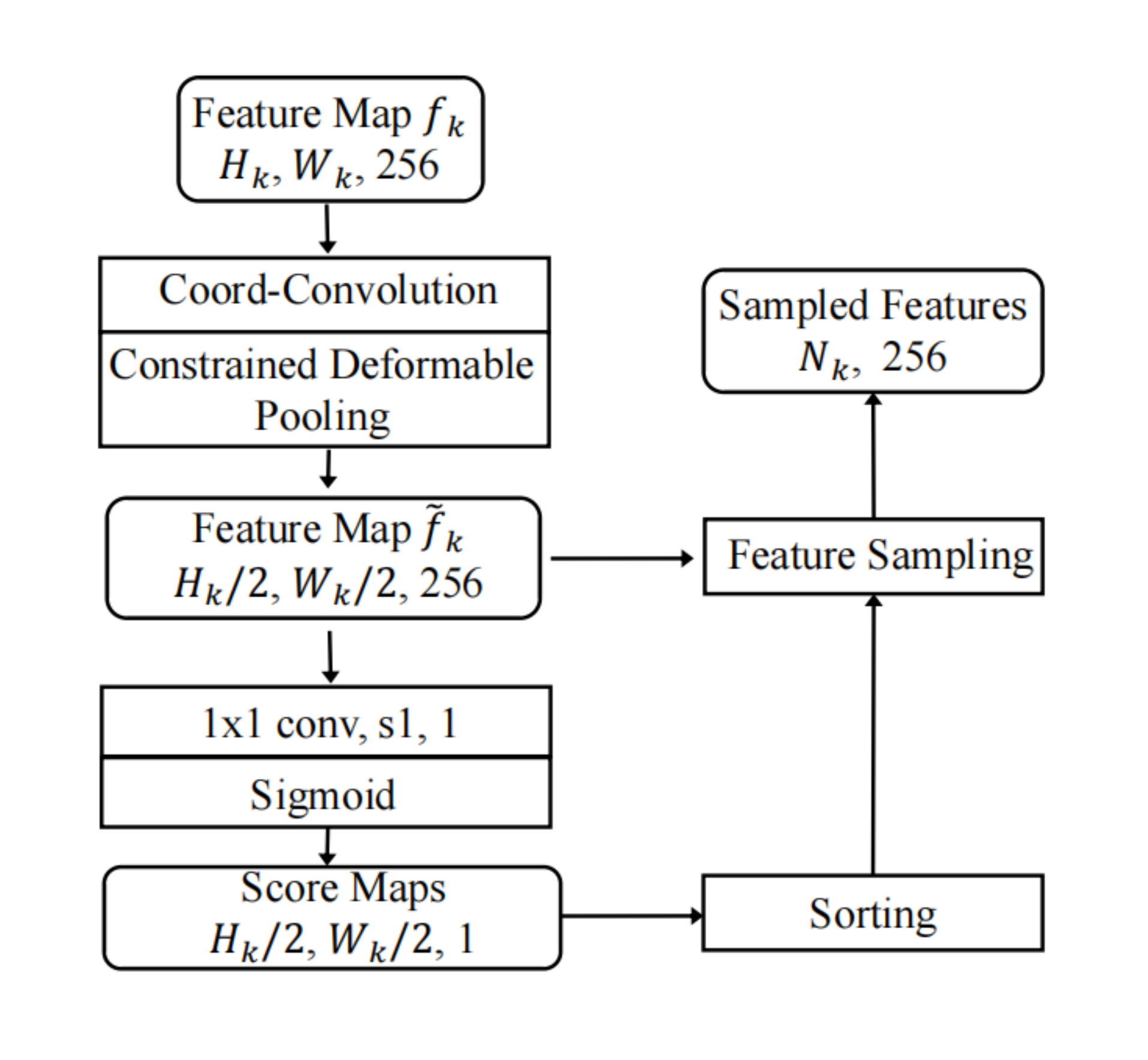}
	\end{center}
	\caption{The pipeline of feature sampling for each feature map $f_k$.
	}
	\label{fig:conf}
\end{figure}

As shown in Fig.~\ref{fig:conf}, each feature map is first fed into a Coord-Convolution layer~\cite{Coord} to involve position information for the incoming presentation in our feature sampling network.
Next, it is down-sampled by a constrained deformable pooling adjusted from~\cite{DCN}.
In our implementation, the predicted offsets are obtained by $\bigtriangleup \mathbf{p}_{ij} = \lambda \cdot \bigtriangleup \mathbf{\widehat{p}}_{ij} \circ (W_k, H_k)$, where $\lambda = Sigmoid(Avg( f_{ij} ))$ is a learnable scaling parameter to modulate the predicted offset and $f_{ij}$ is the feature vector at $(i, j)$. The other symbol definitions are consistent with the original ROI deformable pooling~\cite{DCN}.
Then, a convolution layer with a $1\times1$ kernel size and a Sigmoid function are employed to generate confidence score maps to distinguish representative text regions.
After that, we select the features with top-$N_k$ scores in each scale layer $k$, and gather them into a sequence form with a shape $(\sum_{k}{N_k}, C)$, where $C=256$ is the channel number.

In our feature grouping network, the sampled features are first concatenated with position embeddings.
Then, we adopt four basic transformer encoder layers as those in DETR~\cite{carion2020detr} to model the feature relationship, and implicitly aggregate the features from the same text instance.
Finally, scores and coordinates of rotated bounding boxes are obtained via a text/non-text classification head and a bounding box prediction head, which are composed of full-connected layers and Sigmoid functions.

\begin{figure}[tb]
	\begin{center}
		\includegraphics[width=1.0\linewidth] {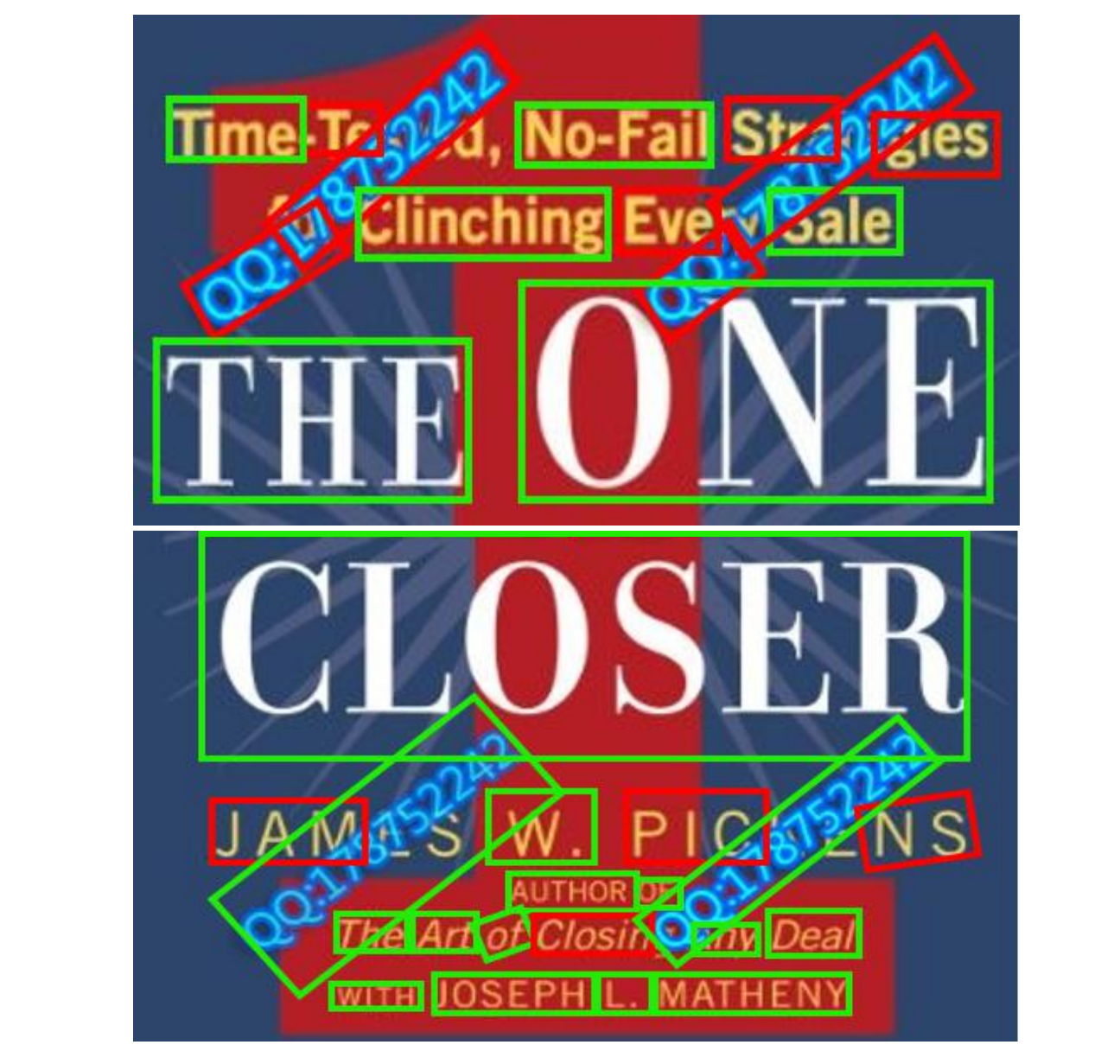}
	\end{center}
	\caption{The bad cases of ``text overlapping" in our method. The red bounding boxes denote the wrong predictions, and the green ones are the right predictions.}
	\label{fig:badcase}
\end{figure}

\subsection{Scale-Invariant GWD Loss}
To regress the coordinates of rotated bounding boxes, we adapt the Gaussian Wasserstein Distance (GWD) loss~\cite{GWD} into a scale-invariant form to better balance the loss weights of text with different scales.
Following the GWD loss, we first convert the rotated bounding box $\mathcal{B}(x,y,h,w,\theta)$ into a 2-D Gaussian distribution representation $\mathcal{N}(\mathbf{m}, \mathbf{\Sigma})$, where $\mathbf{m} = (x,y)$ and $\mathbf{\Sigma}$ is formulated as
\begin{equation}
    \mathbf{\Sigma} = \left( 
    \begin{array}{cc}
         \frac{w}{2}\cos^2\theta+\frac{h}{2}\sin^2\theta&\frac{w-h}{2}\cos\theta\sin\theta  \\
         \frac{w-h}{2}\cos\theta\sin\theta& \frac{w}{2}\sin^2\theta+\frac{h}{2}\cos^2\theta
    \end{array}
    \right)^2.
\end{equation}
Then, we use the Wasserstein distance between two instances to formulate $d^2$ as
\begin{equation}
	d^2 = \parallel \mathbf{m}_1 - \mathbf{m}_2 \parallel_2^2 +
	\textbf{Tr} \left(\mathbf{\Sigma}_1 + \mathbf{\Sigma}_2 - 2(\mathbf{\Sigma}_1^{1/2}\mathbf{\Sigma}_2\mathbf{\Sigma}_1^{1/2})^{1/2} \right).
\end{equation}


Due to the extreme variance of scales, the loss of small text has a negligible influence on the gradient back-propagation compared with the loss of large text.
Hence, we adjust the GWD loss into a scale-invariant form as follows:
\begin{equation}
	\widehat{\mathcal L}_{rbox} = \frac{1}{N_r} \sum_x (1- \frac{1}{\tau + f(d^2({\frac{\widehat u_x}{|\widehat t_x|}}, {\frac{\widehat t_x}{|\widehat t_x|}}))}),
\end{equation}
where $\widehat u_x$ denotes the predicted rotated bounding box, $\widehat t_x$ denotes the target one, and $|\widehat t_x|$ denotes its area.
$N_r$ is the number of bounding boxes after pair-wise matching.
The elements with $\widehat{\quad}$ denote the matched bounding boxes or the target ones after pair-wise matching.
$f(\cdot)$ represents a non-linear function, and $\tau$ is a hyper-parameter to modulate the loss.
According to the GWD loss~\cite{GWD}, we set $f(d^2)=\log{(d^2 + 1)}$ and $\tau=3$.
By normalizing $\widehat u_x$ and $\widehat t_x$ with the area of $\widehat t_x$, we can decrease the negative effect of the scale imbalance.

\begin{figure*}[tb]
	\begin{center}
		\includegraphics[width=1.0\linewidth] {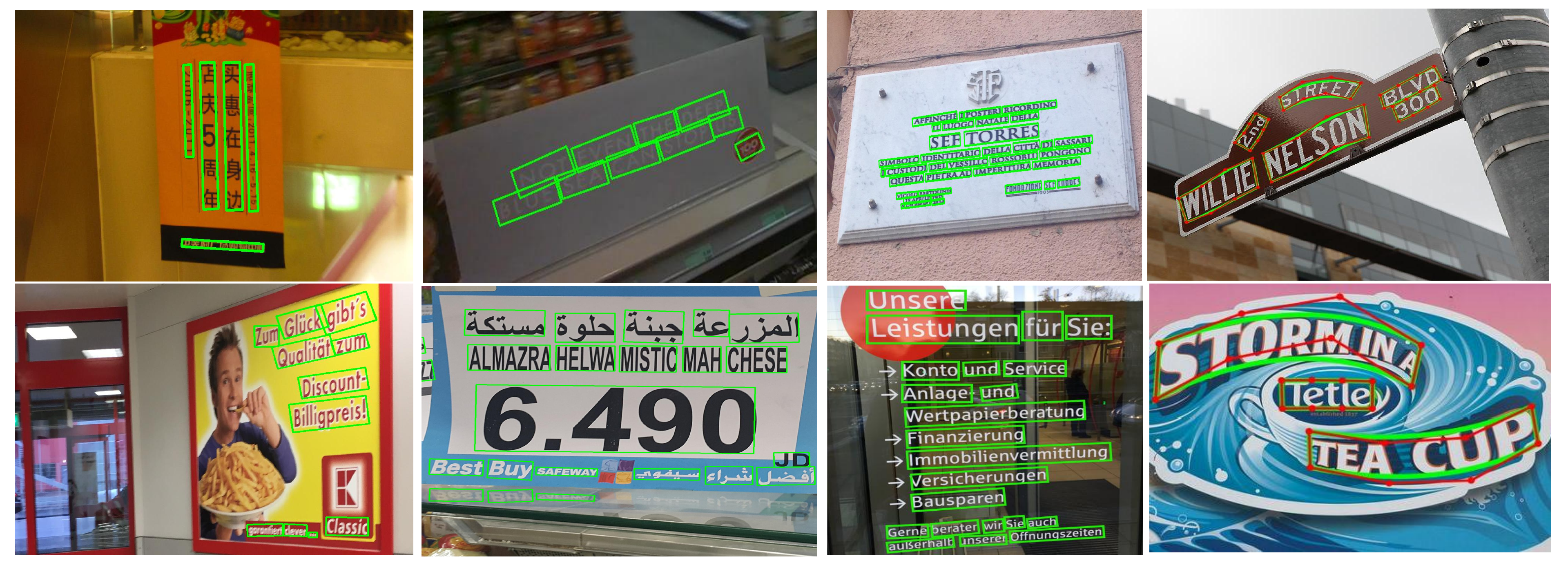}
	\end{center}
	\caption{The qualitative results of our proposed method in different cases, including multi-oriented text, long text, multi-lingual text, low-resolution text, curved text, dense text. For curved text detection, the Bezier curves' control points are drawn in red.
	}
	\label{fig:vis_appendix}
\end{figure*}

\subsection{Training}
In the training period, the data argumentation for training data includes:
(1) Random Rotation, flipping, and perspective transformation;
(2) Color argumentation;
(3) Random cropping.
In addition, both sides of the training images are randomly resized in the range between $640\times640$ and $1680\times1680$ with an interval of 64.
In our loss function, we use $\lambda_c$, $\lambda_d$, and $\lambda_f$ to adjust the influences of different losses.
Specifically, we set $\lambda_c$ to $0.5$ and $\lambda_d$ to 1.
For $\lambda_f$, we initialize it to $1e^{-2}$, and decay it by a factor $0.1$ at the 35th and 45th epoch, respectively.

\subsection{Inference}
In the inference period, we keep the aspect ratio of test images and resize the shorter sides to 768 (for TD500 and MTWI) or 1024 (for others), while the upper limit of the longer sides is 2048.
Moreover, we can easily obtain the detection results without any complex post-processing. By setting a proper threshold, we only keep the predicted boxes with scores higher than the threshold. Specifically, we set it to 0.45 for the IC15 dataset, and 0.5 for others.

\begin{figure}[tb]
	\begin{center}
		\includegraphics[width=1.0\linewidth] {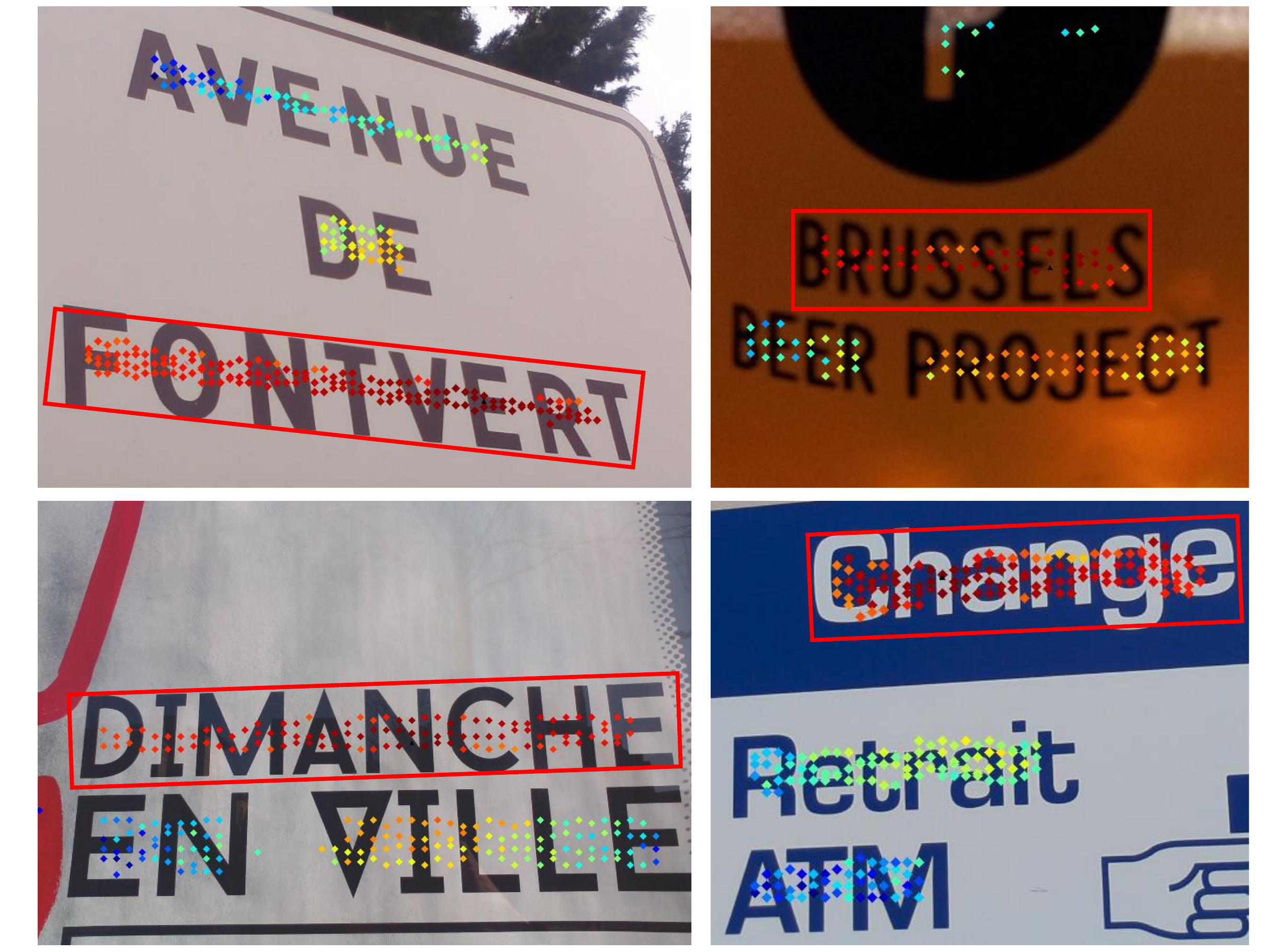}
	\end{center}
	\caption{The visualization of feature sampling and grouping. We visualize the attention weights for one text instance's features in the last transformer layer. The weight value increases from 0 to 1 as the color changes from blue to red. The output feature for the text instance in a red bounding box is mainly aggregated from the inner text point features.
	}
	\label{fig:attn}
\end{figure}

\section{Experiments}

\subsection{Qualitative Results}
As shown in Fig.~\ref{fig:vis_appendix}, we provide more qualitative results for visualization, including multi-oriented text, long text, multi-lingual text, small text, dense text, and curved text.
Moreover, we also provide some bad cases of our method shown in Fig.~\ref{fig:badcase}. The red bounding boxes are the wrong predictions. It is hard for our method to deal with the case of ``text overlapping", because the features of the overlapping text instances are quite complex and tangled. Our feature grouping module sometime fails in these cases.

As shown in Fig.~\ref{fig:attn}, we show the feature grouping results of the predicted rotated bounding boxes in red. We visualize the attention weights for one text instance's features in the last transformer layer.
The weight value increases from 0 to 1 as the color changes from blue to red.
It means that the output features for text instances in red bounding boxes are mainly aggregated from the inner features (red ones).

\subsection{Constrained Deformable Pooling}
To demonstrate the effectiveness of our constrained deformable pooling, we construct an ablation study on the IC15 and the MLT17 datasets.
As shown in Tab.~\ref{tab:pooling}, our constrained deformable pooling outperforms average pooling and the original deformable pooling.
It achieves 89.1\% and 79.5\% f-measure on the IC15 and the MLT17 datasets, respectively.


\subsection{Loss for Rotated Bounding Boxes}
As shown in Tab.~\ref{tab:loss}, we compare the original GWD~\cite{GWD} loss with our proposed scale-invariant form on the IC15 and the MLT17 datasets.
Our scale-invariant GWD loss outperforms the original one by 0.7\% and 0.5\% on the IC15 and the MLT17 datasets.

\begin{table}[t]
	\centering
	\small
	\setlength\tabcolsep{8pt}
	\begin{tabularx}{1\linewidth}{c|c|c|c|c}
		\hline
		\multirow{2}*{Method} &\multicolumn{1}{c}{Sampling} \vline &\multicolumn{3}{c}{F-measure} \\ \cline{3-5}
		~& Number & IC15 & TD500 & MTWI \\
		\hline
			FPN+FC & 64+128+256 & 85.7 & 85.5 &70.6 \\
			FPN+GCN & 64+128+256 & 87.9 	& 87.0 &72.5  \\
			\textbf{Ours (RBox)} & 64+128+256 & \textbf{89.1} & \textbf{88.1} & \textbf{75.2} \\ \hline
	\end{tabularx}
	\caption{The ablation study on feature grouping with non-transformer structures.}
	\label{tab:rebuttal}
\end{table}

\begin{table}[tb]
	\centering
	\small
	\setlength\tabcolsep{5pt}
	\begin{tabularx}{1\linewidth}{c|ccc|ccc}
		\hline
		\multirow{2}*{Methods} & \multicolumn{3}{c}{IC15} \vline &\multicolumn{3}{c}{MLT17 val} \\
		~ & P & R & F & P & R & F\\ 	\hline
		Average Pooling  & 89.5&	87.2&	88.3&86.6	&72.6	&79.0 \\ 
		Deformable Pooling & 89.9&	87.3&	88.6&86.8	&72.8	&79.2 \\ 
		\textbf{Ours (RBox)} &90.9	&87.3	&\textbf{89.1}  &86.8   &73.4 &\textbf{79.5}\\
		\hline
	\end{tabularx}
	\caption{The abalation study on the constrained deformable pooling. ``P", ``R", and ``F" represent Precision, Recall, and F-
measure, respectively.
	}
	\label{tab:pooling}
\end{table}

\begin{table}[tb]
	\centering
	\small
	\setlength\tabcolsep{7pt}
	\begin{tabularx}{1\linewidth}{c|ccc|ccc}
		\hline
		\multirow{2}*{$\widehat{\mathcal L}_{rbox}$} & \multicolumn{3}{c}{IC15} \vline &\multicolumn{3}{c}{MLT17 val} \\
		~ & P & R & F & P & R & F\\ 	\hline
		GWD  &90.2	&86.6	&88.4 	&86.7	&72.6	&79.0  \\ 
	    \textbf{Ours (RBox)} &90.9	&87.3	&\textbf{89.1}  &86.8   &73.4 &\textbf{79.5} \\ 
		\hline
	\end{tabularx}
	\caption{The ablation study on the loss for rotated bounding boxes.
	}
	\label{tab:loss}
\end{table}

\subsection{Compared with Non-Transformer Structure}
To evaluate sampling and grouping with non-transformer methods, we replace our transformer module with GCN~\cite{li2019deepgcns} (FPN+GCN) and FC layers (FPN+FC).
As shown in Tab.~\ref{tab:rebuttal}, these two settings achieve lower f-measure than ours.
This phenomenon validates the effectiveness of our proposed sampling and grouping framework based on transformers.

\end{document}